\documentclass{article}
\usepackage{ijcai26}

\usepackage{times}
\usepackage{soul}
\usepackage{url}
\usepackage[hidelinks]{hyperref}
\usepackage[utf8]{inputenc}
\usepackage[small]{caption}
\usepackage{graphicx}
\usepackage{amsmath}
\usepackage{amsthm}
\usepackage{booktabs}
\usepackage{algorithm}
\usepackage{algorithmic}
\usepackage[switch]{lineno}
\usepackage{amsfonts}
\usepackage{amssymb}
\usepackage{algorithm}
\usepackage{algorithmic}

\title{RefProtoFL: Communication-Efficient Federated Learning via External-Referenced Prototype Alignment}

\iftrue
\author{
Hongyue Wu$^{1,2,3*}$ \and
Hangyu Li$^{1*}$
\and
Guodong Fan$^{1}$ \and
Haoran Zhu$^{1}$ \and
Shizhan Chen$^{1}$ \and
Zhiyong Feng$^{1}$
\affiliations
$^1$College of Intelligence and Computing, Tianjin University\\
$^2$State Key Lab. for Novel Software Technology, Nanjing University\\
$^3$Yunnan Key Lab. of Service Computing, Yunnan University of Finance and Economics\\
\emails
\{hongyue.wu, hangyulee, guodongfan, 2025248021, shizhan, zyfeng\}@tju.edu.cn
}
\fi

\begin{document}
\maketitle

\begin{abstract}
Federated learning (FL) enables collaborative model training without sharing raw data in edge environments, but is constrained by limited communication bandwidth and heterogeneous client data distributions. Prototype-based FL mitigates this issue by exchanging class-wise feature prototypes instead of full model parameters; however, existing methods still suffer from suboptimal generalization under severe communication constraints. In this paper, we propose RefProtoFL, a communication-efficient FL framework that integrates External-Referenced Prototype Alignment (ERPA) for representation consistency with Adaptive Probabilistic Update Dropping (APUD) for communication efficiency. Specifically, we decompose the model into a private backbone and a lightweight shared adapter, and restrict federated communication to the adapter parameters only. To further reduce uplink cost, APUD performs magnitude-aware Top-K sparsification, transmitting only the most significant adapter updates for server-side aggregation. To address representation inconsistency across heterogeneous clients, ERPA leverages a small server-held public dataset to construct external reference prototypes that serve as shared semantic anchors. For classes covered by public data, clients directly align local representations to public-induced prototypes, whereas for uncovered classes, alignment relies on server-aggregated global reference prototypes via weighted averaging. Extensive experiments on standard benchmarks demonstrate that RefProtoFL attains higher classification accuracy than state-of-the-art prototype-based FL baselines.
\end{abstract}
\section{Introduction}
\let\thefootnote\relax\footnotetext{* Equal contribution.}
\let\thefootnote\relax\footnotetext{Code available at: https://github.com/XiaoziLee/RefProtoFL.}
Federated learning provides a principled framework for collaborative model training across distributed clients while preserving data privacy, making it particularly attractive for large-scale edge environments such as mobile and IoT systems. In practice, most federated learning systems are instantiated under the FedAvg paradigm, where clients perform local stochastic optimization and periodically transmit full model updates to a central server for aggregation \cite{mcmahan2017communication}. In these settings, however, practical deployment of FL is often constrained by limited\cite{Chen2024TWC_RobustFL} and unstable communication resources\cite{Tong2025ACM_DRLComm}, as well as highly heterogeneous data distributions across clients\cite{Lu2024JIOT_NonIIDSurvey}. Repeated transmission of full model parameters can therefore incur prohibitive communication costs\cite{zhang2024resourceadaptiveapproachfederatedlearning}, leading to slow convergence and degraded performance, especially when client participation is intermittent\cite{9963723} and data are non-IID\cite{10.5555/3737916.3738181}.

To address communication bottlenecks in federated learning, a range of communication-efficient strategies have been explored. One line of work focuses on reducing the size or frequency of parameter transmission through model compression\cite{malekijoo2021fedzipcompressionframeworkcommunicationefficient}, quantization\cite{wang2025communicationefficientfederatedlearningquantized}, or sparsification\cite{guastella2025sparsyfedsparseadaptivefederated}. While these techniques can significantly lower communication costs, they often require careful tuning and may adversely affect optimization stability under heterogeneous data distributions. Moreover, frequent parameter synchronization, even in compressed form, can remain costly for large models deployed on resource-constrained edge devices.

An alternative line of research exchanges higher-level semantic information instead of raw model parameters. In particular, prototype-based federated learning methods summarize local data distributions using class-wise feature prototypes and communicate these compact representations to the server\cite{tan2022fedproto,fedgh,fedproc}. By decoupling communication cost from model size, such approaches are especially appealing in edge scenarios. However, existing prototype-based methods may still suffer from limited representation consistency across clients, especially when local data are highly non-IID or when certain classes are sparsely represented. As a result, prototype alignment can become unstable, leading to slow convergence and suboptimal generalization.

A central challenge in prototype-based federated learning lies in maintaining consistent semantic representations across distributed clients. When prototypes are computed solely from local data, their semantics can vary substantially due to client-specific data biases, especially under highly non-IID settings\cite{10468591}. This inconsistency may accumulate over communication rounds, causing prototype drift and weakening the effectiveness of prototype alignment as a coordination mechanism\cite{10.1609/aaai.v39i21.34464}.

One natural way to mitigate this issue is to introduce a shared reference that is accessible to all clients. A small amount of public or auxiliary data, when available at the server, can serve as such a reference by providing a common semantic grounding across heterogeneous models. By encouraging local representations to align with reference prototypes induced by shared data, inter-client discrepancies can be reduced without requiring the exchange of raw private data. 

Motivated by the above challenges, we present a communication-efficient federated learning framework that addresses both representation inconsistency and slow convergence in edge environments. The proposed approach consists of two complementary components: External-Referenced Prototype Alignment (ERPA) and Adaptive Probabilistic Update Dropping (APUD). ERPA stabilizes class-wise prototype semantics across heterogeneous clients by incorporating shared external references, thereby mitigating prototype drift induced by client-specific data biases. In parallel, APUD improves communication efficiency by selectively transmitting informative model updates under constrained uplink budgets, allowing the server to aggregate compact yet effective optimization signals. 

The main contributions of this work are summarized as follows:
\begin{itemize}
    \item We propose \textbf{RefProtoFL}, a unified federated learning framework that jointly integrates \textbf{External-Referenced Prototype Alignment (ERPA)} and \textbf{Adaptive Magnitude-based Update Dropping (APUD)} to achieve robust representation alignment and communication-efficient training under non-IID data distributions in edge environments.
    
    \item We design a \textbf{reference-guided prototype alignment mechanism} that exploits a small server-held public dataset to construct class-wise external anchors, and further aggregates client-side prototypes for classes absent from the public data, enabling consistent representation learning without exposing private samples.
    
    \item To substantially reduce uplink communication overhead, we introduce a \textbf{magnitude-aware Top-$K$ adapter update dropping strategy} that selectively transmits only the most significant adapter parameters while preserving convergence behavior and model performance.
    
    \item Extensive experimental results on multiple benchmarks demonstrate that the proposed method consistently outperforms existing prototype-based federated learning approaches in both communication-constrained and highly heterogeneous settings.
\end{itemize}

\section{Related Work}
\subsection{Efficient federated optimization}
Federated learning in edge environments is fundamentally constrained by limited communication bandwidth and heterogeneous data distributions. Early efforts address heterogeneity from an optimization perspective. FedProx introduces a proximal regularization term to stabilize local training under non-IID data and partial client participation \cite{li2020federated}. FedNova further identifies objective inconsistency caused by heterogeneous local update steps and proposes normalized aggregation to restore unbiased optimization \cite{wang2020tackling}. Although these methods improve convergence robustness, they still rely on frequent transmission of model parameters or gradients, which remains costly for bandwidth-constrained edge systems.

\subsection{Prototype-based federated learning}
To reduce communication overhead, prototype-based methods replace parameter exchange with compact class-wise feature representations. FedProto communicates local prototypes to the server for aggregation and uses global prototypes to regularize local learning, enabling collaboration across heterogeneous clients with reduced communication cost \cite{tan2022fedproto}. FedPCL further leverages prototype-wise supervised contrastive learning in a pre-trained model setting, demonstrating that prototypes can serve as effective carriers of semantic information in federated training \cite{tan2022federated}. However, when prototypes are computed solely from private local data, severe non-IID distributions can induce semantic inconsistency across clients, leading to prototype drift and unstable alignment.

\subsection{External references for federated learning}
Several works incorporate shared external information to mitigate representation inconsistency in federated learning. Virtual Homogeneity Learning introduces shared virtual data to calibrate feature distributions across heterogeneous clients, improving robustness under data heterogeneity \cite{tang2022virtual}. FedProtoKD combines private-data prototypes with public-data knowledge distillation to enhance semantic alignment and address degradation caused by naive prototype aggregation \cite{hossen2025fedprotokd}. While effective, these approaches typically employ external data as an auxiliary signal and do not explicitly anchor prototype semantics to a shared reference, nor do they directly address strict uplink bandwidth constraints.

\section{Problem Statement}
A federated learning system consists of a central server and a set of \(K\) clients indexed by \(\mathcal{K}=\{1,\dots,K\}\).
Each client \(k\) holds a local private dataset \(\mathcal{D}_k=\{(x_i,y_i)\}_{i=1}^{|\mathcal{D}_k|}\), which cannot be shared with the server or other clients due to privacy constraints.
The data distributions across clients are generally non-identically distributed (non-IID),
\begin{equation}
\mathcal{D}_k \sim \mathcal{P}_k, \qquad \mathcal{P}_k \neq \mathcal{P}_{k'}, \ \forall k \neq k'.
\label{eq:1}
\end{equation}

The goal of FL is to collaboratively learn a global model parameterized by \(\theta\) by minimizing the weighted empirical risk over all participating clients,
\begin{equation}
\min_{\theta} \; \mathcal{L}(\theta)
=
\sum_{k=1}^{K} \frac{|\mathcal{D}_k|}{\sum_{j=1}^{K} |\mathcal{D}_j|}
\;\mathbb{E}_{(x,y)\sim \mathcal{D}_k}
\big[
\ell(f(x;\theta),y)
\big],
\label{eq:2}
\end{equation}
where \(f(\cdot;\theta)\) denotes the global model and \(\ell(\cdot,\cdot)\) is the task loss.

Training is conducted in communication rounds.
At each round \(t\), the server broadcasts the current global model to a subset of clients, each selected client performs local optimization on its private data, and the server aggregates the uploaded updates to obtain the next global model.
In practical edge environments, FL is often subject to strict communication constraints and pronounced data heterogeneity, which together pose significant challenges to efficient and robust global model convergence.

\section{Methodology}
\subsection{Adaptive Probabilistic Update Dropping}
Let \( \theta^{a}\in\mathbb{R}^{d} \) denote the parameters of the shared adapter module, which is the only component participating in federated communication via APUD. At communication round \( t \), the server broadcasts the global adapter parameters \( \theta^{a,t} \) to a subset of clients. Each client \( k \) performs local training and obtains updated adapter parameters \( \theta^{a,t}_{k} \).

\begin{algorithm}[t]
\caption{APUD on client $k$ at round $t$}
\label{alg:apud_client}
\begin{algorithmic}[1]
\REQUIRE Global adapter $\theta^{a,t}\in\mathbb{R}^d$; local updated adapter $\theta^{a,t}_k$; budget $K\ll d$
\ENSURE Masked adapter $\theta^{a,t}_k\odot M^t_k$ and mask $M^t_k\in\{0,1\}^d$
\STATE $u^t_k \leftarrow \left|\theta^{a,t}_k - \theta^{a,t}\right|$ \COMMENT{element-wise magnitude}
\STATE $\mathcal{S}^t_k \leftarrow \operatorname{arg\,top}_K\{u^t_{k,i}\}_{i=1}^{d}$
\STATE $M^t_k \leftarrow \mathbf{0}\in\{0,1\}^d$
\FOR{each $i \in \mathcal{S}^t_k$}
    \STATE $M^t_{k,i} \leftarrow 1$
\ENDFOR
\STATE Transmit $\left(\theta^{a,t}_k\odot M^t_k,\; M^t_k\right)$ to server
\end{algorithmic}
\end{algorithm}

\begin{algorithm}[t]
\caption{APUD server aggregation at round $t$}
\label{alg:apud_server}
\begin{algorithmic}[1]
\REQUIRE Global adapter $\theta^{a,t}\in\mathbb{R}^d$; received $\{(\theta^{a,t}_k\odot M^t_k, M^t_k)\}_{k\in\mathcal{K}^t}$; data sizes $\{|\mathcal{D}_k|\}$
\ENSURE Updated global adapter $\theta^{a,t+1}$
\FOR{$i=1$ to $d$}
    \STATE $\mathcal{K}^t_i \leftarrow \{k \mid M^t_{k,i}=1\}$
    \IF{$\mathcal{K}^t_i \neq \varnothing$}
        \STATE $Z \leftarrow \sum_{j\in\mathcal{K}^t_i} |\mathcal{D}_j|$
        \STATE $\theta^{a,t+1}_i \leftarrow \sum_{k\in\mathcal{K}^t_i}\left(\frac{|\mathcal{D}_k|}{Z}\right)\theta^{a,t}_{k,i}$
    \ELSE
        \STATE $\theta^{a,t+1}_i \leftarrow \theta^{a,t}_i$
    \ENDIF
\ENDFOR
\end{algorithmic}
\end{algorithm}

\subsubsection{Update Magnitude Estimation}
To determine the importance of each adapter parameter, client \( k \) computes the element-wise update magnitude vector:
\begin{equation}
u^{t}_{k} = \left| \theta^{a,t}_{k} - \theta^{a,t} \right|,
\label{eq:3}
\end{equation}
where the absolute value is applied element-wise.
The vector \( u^{t}_{k} \) is used only for parameter selection, not for aggregation. 
\subsubsection{Top-K Parameter Selection}
Given a communication budget \( K \ll d \), APUD selects the indices corresponding to the \( K \) largest-magnitude updates in \( u^{t}_{k} \). Formally, define
\begin{equation}
\mathcal{S}^{t}_{k} = \operatorname{arg\,top}_K \left\{ u^{t}_{k,i} \mid i = 1, \dots, d \right\},
\label{eq:4}
\end{equation}
where \( \operatorname{arg\,top}_K(\cdot) \) returns the indices of the \( K \) largest entries by magnitude.

A binary selection mask \( M^{t}_{k} \in \{0,1\}^{d} \) is then constructed as:
\begin{equation}
M^{t}_{k,i} =
\begin{cases}
1, & i \in \mathcal{S}^{t}_{k}, \\
0, & \text{otherwise}.
\end{cases}
\label{eq:5}
\end{equation}
Only the masked adapter parameters \( \theta^{a,t}_{k} \odot M^{t}_{k} \), together with the mask \( M^{t}_{k} \), are transmitted to the server.

\subsubsection{Federated Aggregation}
Let \( \mathcal{D}_k \) denote the local dataset on client \( k \), and \( |\mathcal{D}_k| \) its cardinality.
At communication round \( t \), each client transmits a masked subset of adapter parameters according to the selection mask \( M_k^t \).

For each adapter parameter index \( i \in \{1,\dots,d\} \), we define the set of clients that transmit the \(i\)-th parameter at round \( t \) as
\begin{equation}
\mathcal{K}_i^{t}
=
\left\{\, k \;\middle|\; M^{t}_{k,i} = 1 \,\right\}.
\label{eq:6}
\end{equation}

The server updates the global adapter parameters in an element-wise manner. Specifically, the \(i\)-th parameter is updated as:
\begin{equation}
\theta^{a,t+1}_i =
\begin{cases}
\displaystyle
\sum_{k \in \mathcal{K}_i^{t}} 
\underbrace{\frac{|\mathcal{D}_k|}{\sum_{j \in \mathcal{K}_i^{t}} |\mathcal{D}_j|}}_{\text{weight}} 
\theta^{a,t}_{k,i},
& \text{if } \mathcal{K}_i^{t} \neq \varnothing, \\[8pt]
\theta^{a,t}_i,
& \text{if } \mathcal{K}_i^{t} = \varnothing.
\end{cases}
\label{eq:7}
\end{equation}

That is, only adapter parameters that are selected and transmitted by at least one client participate in aggregation, where each parameter is aggregated using a data-size-weighted average over the corresponding transmitting clients. Parameters that are not transmitted by any client at round \( t \) remain unchanged. Client-side masking is summarized in Algorithm.\ref{alg:apud_client}, and the corresponding server-side masked aggregation is given in Algorithm.\ref{alg:apud_server}.

\subsection{External-Referenced Prototype Alignment}
At communication round \(t\), the server broadcasts the global adapter parameters \( \theta^{a,t} \).
For clarity, we partition them into the feature-adaptation part and the classifier head:
\begin{equation}
\theta^{a,t} = \big(\theta^{f,t},\,\theta^{c,t}\big),
\label{eq:split_theta_a}
\end{equation}
where \(\theta^{f,t}\) denotes the feature transformation parameters and \(\theta^{c,t}\) denotes the classifier-head parameters.

\begin{algorithm}[t]
\caption{ERPA client prototype computation on client $k$ at round $t$}
\label{alg:erpa_client_proto}
\begin{algorithmic}[1]
\REQUIRE Classes $\mathcal{C}$; public dataset $\mathcal{D}^{\mathrm{pub}}$ with $\mathcal{D}^{\mathrm{pub}}_c$; local dataset $\mathcal{D}_k$ with $\mathcal{D}_{k,c}$; broadcast adapter $\theta^{a,t}=(\theta^{f,t},\theta^{c,t})$; client-private backbone $\theta^{b,t}_k$
\ENSURE Uploaded public prototypes $\{p^{\mathrm{pub},t}_{k,c}\}_{|\mathcal{D}^{\mathrm{pub}}_c|>0}$ and local prototypes $\{(p^{t}_{k,c},|\mathcal{D}_{k,c}|)\}_{|\mathcal{D}^{\mathrm{pub}}_c|=0,|\mathcal{D}_{k,c}|>0}$
\FOR{each class $c\in\mathcal{C}$}
    \IF{$|\mathcal{D}^{\mathrm{pub}}_c|>0$}
        \STATE \parbox[t]{0.96\linewidth}{$\displaystyle
        p^{\mathrm{pub},t}_{k,c}\leftarrow
        \frac{1}{|\mathcal{D}^{\mathrm{pub}}_c|}
        \sum_{x\in\mathcal{D}^{\mathrm{pub}}_c}
        F\!\left(x;\theta^{b,t}_{k},\theta^{f,t}\right)
        $}
        \STATE Upload $p^{\mathrm{pub},t}_{k,c}$.
    \ELSIF{$|\mathcal{D}^{\mathrm{pub}}_c|=0 \ \land\ |\mathcal{D}_{k,c}|>0$}
        \STATE \parbox[t]{0.96\linewidth}{$\displaystyle
        p^{t}_{k,c}\leftarrow
        \frac{1}{|\mathcal{D}_{k,c}|}
        \sum_{x\in\mathcal{D}_{k,c}}
        F\!\left(x;\theta^{b,t}_{k},\theta^{f,t}\right)
        $}
        \STATE Upload $\big(p^{t}_{k,c},\,|\mathcal{D}_{k,c}|\big)$.
    \ENDIF
\ENDFOR
\end{algorithmic}
\end{algorithm}

\begin{algorithm}[t]
\caption{ERPA server prototype aggregation and broadcast at round $t$}
\label{alg:erpa_server_proto}
\begin{algorithmic}[1]
\REQUIRE Classes $\mathcal{C}$; selected clients $\mathcal{S}^t$; public availability $\{|\mathcal{D}^{\mathrm{pub}}_c|\}$; received $\{p^{\mathrm{pub},t}_{k,c}\}$ and $\{(p^{t}_{k,c},|\mathcal{D}_{k,c}|)\}$
\ENSURE External-reference prototypes $\{p^{\mathrm{ext},t}_c\}_{\delta_c=0}$; global prototypes $\{p^{\mathrm{g},t}_c\}_{\delta_c=1,\mathcal{U}^t_c\neq\varnothing}$
\FOR{each class $c\in\mathcal{C}$}
    \STATE $\delta_c \leftarrow \mathbb{I}\!\left(|\mathcal{D}^{\mathrm{pub}}_c|=0\right)$
    \IF{$\delta_c=0$}
        \STATE \parbox[t]{0.96\linewidth}{$\displaystyle
        p^{\mathrm{ext},t}_c \leftarrow
        \frac{1}{|\mathcal{S}^t|}
        \sum_{k\in\mathcal{S}^t} p^{\mathrm{pub},t}_{k,c}
        $}
    \ELSE
        \STATE $\mathcal{U}^{t}_c \leftarrow \left\{\, k \in \mathcal{S}^t \ \middle|\  |\mathcal{D}_{k,c}|>0 \,\right\}$
        \IF{$\mathcal{U}^{t}_c\neq\varnothing$}
            \STATE \parbox[t]{0.96\linewidth}{$\displaystyle
            p^{\mathrm{g},t}_c \leftarrow
            \sum_{k\in\mathcal{U}^{t}_c}
            \frac{|\mathcal{D}_{k,c}|}{\sum_{j\in\mathcal{U}^{t}_c}|\mathcal{D}_{j,c}|}
            \, p^{t}_{k,c}
            $}
        \ENDIF
    \ENDIF
\ENDFOR
\STATE Broadcast $\{p^{\mathrm{ext},t}_c\}_{c:\delta_c=0}$ and $\{p^{\mathrm{g},t}_c\}_{c:\delta_c=1,\mathcal{U}^{t}_c\neq\varnothing}$.
\end{algorithmic}
\end{algorithm}

\begin{algorithm}[t]
\caption{ERPA anchor-based local optimization on client $k$ at round $t$}
\label{alg:erpa_client_train}
\begin{algorithmic}[1]
\REQUIRE Classes $\mathcal{C}$; local dataset $\mathcal{D}_k$ with $\mathcal{D}_{k,c}$; received prototypes $\{p^{\mathrm{ext},t}_c\}_{\delta_c=0}$ and $\{p^{\mathrm{g},t}_c\}_{\delta_c=1,\mathcal{U}^t_c\neq\varnothing}$; indicators $\{\delta_c\}$; initialization $(\theta^{b,t}_k,\theta^{a,t}=(\theta^{f,t},\theta^{c,t}))$; trade-off $\lambda$
\ENSURE Updated local adapter $\theta^{a,t}_k$
\FOR{each class $c\in\mathcal{C}$}
    \STATE $\displaystyle a^{t}_{k,c}\leftarrow (1-\delta_c)\,p^{\mathrm{ext},t}_c + \delta_c\,p^{\mathrm{g},t}_c$
\ENDFOR
\STATE Initialize $(\theta^{b}_{k},\theta^{a}) \leftarrow (\theta^{b,t}_{k},\theta^{a,t})$.
\STATE Optimize $(\theta^{b}_{k},\theta^{a})$ on $\mathcal{D}_k$ by minimizing
$\mathcal{L}_k=\mathcal{L}^{\mathrm{ce}}_{k}+\lambda\,\mathcal{L}^{\mathrm{proto}}_{k}$
\STATE Output updated adapter as $\theta^{a,t}_k$; keep $\theta^{b}_{k}$ private.
\end{algorithmic}
\end{algorithm}

\begin{algorithm}[t]
\caption{RefProtoFL}
\label{alg:refprotofl}
\begin{algorithmic}[1]
\REQUIRE Total communication rounds $T$; classes $\mathcal{C}$; public dataset $\mathcal{D}^{\mathrm{pub}}$; client datasets $\{\mathcal{D}_k\}_{k=1}^{N}$; initial global adapter $\theta^{a,0}=(\theta^{f,0},\theta^{c,0})$; APUD budget $K\ll d$; trade-off $\lambda$
\ENSURE Final global adapter $\theta^{a,T}$

\STATE \textbf{Server:} initialize $\theta^{a,0}$.
\FOR{$t=0$ to $T-1$}
    \STATE \textbf{Server:} broadcast $\theta^{a,t}$ to all $k\in\mathcal{S}^t$ and provide $\mathcal{D}^{\mathrm{pub}}$.

    \FOR{each client $k\in\mathcal{S}^t$}
        \STATE Run Algorithm.\ref{alg:erpa_client_proto} and upload required prototypes.
    \ENDFOR

    \STATE Run Algorithm.\ref{alg:erpa_server_proto} to obtain $\{p^{\mathrm{ext},t}_c\}_{c:\delta_c=0}$ and $\{p^{\mathrm{g},t}_c\}_{c:\delta_c=1,\mathcal{U}^t_c\neq\varnothing}$.
    \STATE Broadcast $\{p^{\mathrm{ext},t}_c\}$ and $\{p^{\mathrm{g},t}_c\}$ to all $k\in\mathcal{S}^t$.

    \FOR{each client $k\in\mathcal{S}^t$}
        \STATE Run Algorithm.\ref{alg:erpa_client_train} to obtain updated local adapter $\theta^{a,t}_k$.
    \ENDFOR

    \FOR{each client $k\in\mathcal{S}^t$}
        \STATE Apply Algorithm.\ref{alg:apud_client} to produce $\left(\theta^{a,t}_k\odot M^t_k,\; M^t_k\right)$ and transmit to server.
    \ENDFOR

    \STATE Aggregate received masked adapters via Algorithm.\ref{alg:apud_server} to obtain $\theta^{a,t+1}$.
\ENDFOR
\STATE \textbf{Return} $\theta^{a,T}$.
\end{algorithmic}
\end{algorithm}

On each client \(k\), the backbone parameters \( \theta^{b}_{k} \) are optimized locally and are kept private, without being transmitted to the server.
At communication round \(t\), the server broadcasts the global adapter parameters \( \theta^{a,t} \), while each participating client \(k\in\mathcal{S}^t\) maintains its local backbone \( \theta^{b,t}_{k} \), where \(\mathcal{S}^t\) denotes the set of clients selected at round \(t\).
We define the client-specific embedding function as
\begin{equation}
z^{t}_{k}(x) = F\!\left(x;\theta^{b,t}_{k},\theta^{f,t}\right)\in\mathbb{R}^{m},
\label{eq:8}
\end{equation}
which excludes the classifier head \(\theta^{c,t}\). In prototype learning, a class prototype is defined as the empirical mean of representations in the embedding space.

\subsubsection{External-reference prototypes}
The server maintains a small public dataset \( \mathcal{D}^{\mathrm{pub}} \) and provides it to clients.
Let \( \mathcal{D}^{\mathrm{pub}}_c=\{x\in\mathcal{D}^{\mathrm{pub}}\mid y(x)=c\} \).
Since the representation \(F(\cdot;\theta^{b,t}_{k},\theta^{f,t})\) depends on client-private backbones, the server does not directly embed public samples.
Instead, upon receiving \( \theta^{a,t} \), each participating client \(k\in\mathcal{S}^t\) computes a public prototype for each class \(c\) with \( |\mathcal{D}^{\mathrm{pub}}_c|>0 \) using its local backbone together with the broadcast adapter:
\begin{equation}
p^{\mathrm{pub},t}_{k,c}
=
\frac{1}{|\mathcal{D}^{\mathrm{pub}}_c|}
\sum_{x\in\mathcal{D}^{\mathrm{pub}}_c}
F\!\left(x;\theta^{b,t}_{k},\theta^{f,t}\right),
\quad \text{for }|\mathcal{D}^{\mathrm{pub}}_c|>0.
\label{eq:10}
\end{equation}
Here, \( \mathcal{D}^{\mathrm{pub}} \) serves as a shared anchor set that enables the construction of cross-client external-reference prototypes through federated aggregation.

\subsubsection{Local prototypes for classes missing in public data}
On client \(k\), let \( \mathcal{D}_{k,c}=\{x\in\mathcal{D}_{k}\mid y(x)=c\} \).
For classes with \( |\mathcal{D}^{\mathrm{pub}}_c|=0 \) and \( |\mathcal{D}_{k,c}|>0 \), client \(k\) computes a local prototype
\begin{equation}
p^{t}_{k,c}
=
\frac{1}{|\mathcal{D}_{k,c}|}
\sum_{x\in\mathcal{D}_{k,c}}
F\!\left(x;\theta^{b,t}_{k},\theta^{f,t}\right).
\label{eq:11}
\end{equation}
Only these class-level vectors \( p^{t}_{k,c} \) are transmitted to the server; no backbone parameters are exchanged.

\subsubsection{Federated aggregation of prototypes}
Define the public-availability indicator
\begin{equation}
\delta_c=\mathbb{I}\!\left(|\mathcal{D}^{\mathrm{pub}}_c|=0\right).
\label{eq:12}
\end{equation}
For each class \(c\), define the set of participating clients that upload its prototype at round \(t\):
\begin{equation}
\mathcal{U}^{t}_c
=
\left\{\, k \in \mathcal{S}^t \;\middle|\; \delta_c=1,\; |\mathcal{D}_{k,c}|>0 \,\right\}.
\label{eq:13}
\end{equation}

For each class \(c\) with \(\delta_c=0\), the server aggregates an external-reference prototype via uniform aggregation over the uploaded public prototypes:
\begin{equation}
p^{\mathrm{ext},t}_c
=
\frac{1}{|\mathcal{S}^t|}
\sum_{k\in\mathcal{S}^t}
p^{\mathrm{pub},t}_{k,c},
\qquad \text{if } \delta_c = 0.
\label{eq:9}
\end{equation}

The server broadcasts \( \{p^{\mathrm{ext},t}_c\}_{c:\delta_c=0} \) to clients.

The server aggregates a global prototype for missing-public classes as
\begin{equation}
p^{\mathrm{g},t}_c
=
\begin{cases}
\displaystyle
\sum_{k\in\mathcal{U}^{t}_c}
\frac{|\mathcal{D}_{k,c}|}{\sum_{j\in\mathcal{U}^{t}_c}|\mathcal{D}_{j,c}|}
\, p^{t}_{k,c},
& \text{if } \mathcal{U}^{t}_c\neq\varnothing,\\[10pt]
\text{undefined},
& \text{otherwise}.
\end{cases}
\label{eq:14}
\end{equation}
The server broadcasts \( \{p^{\mathrm{g},t}_c\}_{c:\delta_c=1,\,\mathcal{U}^{t}_c\neq\varnothing} \) to clients.

\subsubsection{Prototype alignment objective}
For client \(k\), define the class anchor at round \(t\) as
\begin{equation}
a^{t}_{k,c}
=
(1-\delta_c)\,p^{\mathrm{ext},t}_c
+
\delta_c\,p^{\mathrm{g},t}_c.
\label{eq:15}
\end{equation}

During local optimization at round \(t\), client \(k\) updates \((\theta^{b}_{k},\theta^{a})\) starting from \((\theta^{b,t}_{k},\theta^{a,t})\) by minimizing a prototype alignment loss that encourages sample representations to concentrate around the corresponding class anchors,
\begin{equation}
\mathcal{L}^{\mathrm{proto}}_{k}
=
\sum_{c\in\mathcal{C}}
\sum_{x\in\mathcal{D}_{k,c}}
\left\| F\!\left(x;\theta^{b}_{k},\theta^{f}\right)-a^{t}_{k,c} \right\|_2^{2}.
\label{eq:16}
\end{equation}
In addition to prototype alignment, the client is also optimized with the standard classification objective to preserve discriminative performance.
Specifically, we adopt the cross-entropy loss over the local labeled data,
\begin{equation}
\mathcal{L}^{\mathrm{ce}}_{k}
=
\sum_{(x,y)\in\mathcal{D}_{k}}
\ell_{\mathrm{ce}}\!\left(
g\!\left(F(x;\theta^{b}_{k},\theta^{f});\,\theta^{c}\right),
y
\right),
\label{eq:17}
\end{equation}
where \(g(\cdot;\theta^{c})\) denotes the classifier head parameterized by \(\theta^{c}\), and \(\ell_{\mathrm{ce}}(\cdot,\cdot)\) is the cross-entropy loss.
The overall local training objective for client \(k\) at round \(t\) is then given by
\begin{equation}
\mathcal{L}_{k}
=
\mathcal{L}^{\mathrm{ce}}_{k}
+
\lambda\,\mathcal{L}^{\mathrm{proto}}_{k},
\label{eq:18}
\end{equation}
where \(\lambda\) controls the trade-off between discriminative supervision and prototype-based representation alignment.

Algorithm \ref{alg:refprotofl} shows the complete RefProtoFL process.

\section{Experiments}
\subsection{Experimental Setups}
\subsubsection{Datasets}
\begin{figure}[t]
  \centering
  \includegraphics[width=\columnwidth]{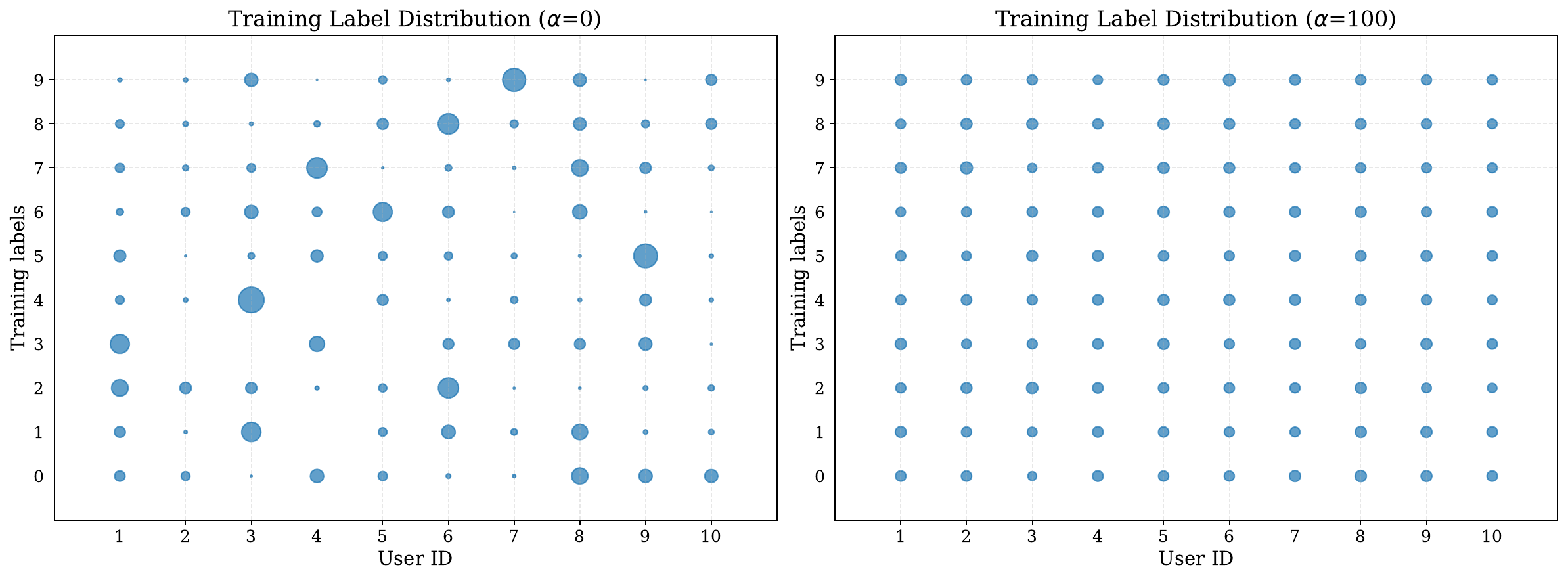}
  \caption{Visualization of data partitioning in the CIFAR10 dataset.}
  \label{fig:CIFAR10_distribution}
\end{figure}

We evaluate the proposed method and several state-of-the-art baselines on four widely-used image classification benchmarks: CIFAR10, CIFAR100\cite{krizhevsky2009learning}, FashionMNIST\cite{xiao2017fashion}, and MNIST\cite{mnist}.
CIFAR10 consists of 60,000 color images of size $32\times32$ from 10 classes, with 6,000 images per class.  
CIFAR100 contains 60,000 images of the same resolution but with 100 fine-grained classes, each having 600 images.  
Both datasets are split into 50,000 training samples and 10,000 test samples.
FashionMNIST and MNIST are grayscale image datasets of size $28\times28$, each containing 70,000 samples across 10 categories.  
They are divided into 60,000 training samples and 10,000 test samples.
To simulate non-IID data distributions, the training data are partitioned among $N=10$ clients using a Dirichlet distribution $\mathrm{Dir(\alpha)}$ over class proportions, with the concentration parameter set to $\alpha=0.5$ and $\alpha=100$. 
Figure \ref{fig:CIFAR10_distribution} illustrates the specific dataset partitioning scheme, using CIFAR10 as an example.

\subsubsection{Training Details}
For all experiments, we adopt ResNet18\cite{he2016deep} architecture as the backbone network, since standard batch normalization may degrade performance in federated learning with heterogeneous client data. 
Local model optimization is performed using SGD with a learning rate $\eta=0.01$, momentum $0.9$, and weight decay $1\times10^{-4}$.
The local batch size is set to $B=64$.  
In each communication round, clients conduct $E=1$ epoch of local training, and the total number of communication rounds is set to $T=50$. All experiments are implemented in PyTorch and executed on a workstation equipped with an NVIDIA GeForce RTX 4060 Ti GPU.

\subsubsection{Baselines}
We compare our method with several representative federated learning baselines, including FedAvg\cite{mcmahan2017communication}, FedProx\cite{li2020federated}, FedNova\cite{wang2020tackling}, FedProto\cite{tan2022fedproto}, VHL\cite{tang2022virtual}, FedGH\cite{fedgh}, and FAST\cite{li2025fast}.

\subsubsection{Benchmarks}
We evaluate federated learning methods using Top-1 accuracy and the total communication cost, measured as the number of transmitted model parameters.

\subsection{Main Results}
\subsubsection{Top-1 accuracy}
Table \ref{tab:full_benchmark} reports the Top-1 accuracy of RefProtoFL and seven representative baselines across four widely-used image classification benchmarks under two Dirichlet partition settings, \(\alpha\in\{0.5,100\}\). Overall, RefProtoFL achieves the best averaged performance (\(60.63\%\)) across all datasets and heterogeneity regimes, surpassing the strongest baseline FedProto (\(60.11\%\)) as well as optimization-based methods (FedAvg/FedProx/FedNova) and personalized or heterogeneous FL baselines (VHL/FedGH/FAST). This consistent advantage suggests that RefProtoFL is not tailored to a specific dataset or a single heterogeneity condition, but instead provides robust improvements across diverse data complexity levels and distribution shifts.

\paragraph{Performance under high heterogeneity (\(\alpha=0.5\)).}
When the data are highly non-IID, RefProtoFL delivers the most consistent gains and ranks first on all four datasets. Compared with the best competing method in this regime, RefProtoFL improves accuracy by \(+0.28\%\) on MNIST (88.46 vs.\ 88.18), \(+0.39\%\) on FashionMNIST (74.21 vs.\ 73.82), \(+1.18\%\) on CIFAR-10 (45.51 vs.\ 44.33), and \(+0.22\%\) on CIFAR-100 (16.50 vs.\ 16.28). RefProtoFL consistently outperforms FedAvg/FedProx/FedNova, indicating that purely optimization-centric corrections are insufficient to fully address the strong distribution mismatch in this setting. The strong results under \(\alpha=0.5\) highlight the effectiveness of RefProtoFL in aggregating knowledge across skewed clients and in mitigating the degradation commonly observed under extreme label imbalance.

\paragraph{Performance under low heterogeneity (\(\alpha=100\)).}
Under a relatively mild heterogeneity setting, RefProtoFL remains highly competitive and achieves the best results on the two more difficult benchmarks, i.e., CIFAR-10 and CIFAR-100. Notably, it reaches 56.36 on CIFAR-10, slightly exceeding FedProto (56.05), and attains 20.02 on CIFAR-100, improving over FedProto by a clear margin (20.02 vs.\ 18.42, \(+1.60\%\)). This suggests that, even when the data distribution is closer to IID, RefProtoFL can still enhance representation learning and global generalization on complex datasets. On the other hand, RefProtoFL is marginally outperformed by FedProto on MNIST (98.25 vs.\ 98.33, \(-0.08\%\)) and FashionMNIST (85.76 vs.\ 85.96, \(-0.20\%\)). A plausible explanation is that these two datasets are relatively simple and already approach a performance ceiling in the low-heterogeneity regime, leaving limited room for further improvement. In such cases, the additional mechanisms introduced by RefProtoFL may yield diminishing returns, and small fluctuations can occur due to optimization noise rather than systematic shortcomings. Importantly, the gaps are minor, indicating that RefProtoFL does not trade off performance in near-IID scenarios while still offering clear benefits on harder tasks.

\subsubsection{Communication Cost}
We analyze the communication overhead of different federated learning methods in terms of the amount of information transmitted between clients and the server per communication round.
Conventional baselines such as FedAvg, FedProx, FedNova, VHL, and FAST adopt a full-model aggregation paradigm, where each participating client uploads the complete set of model parameters at every round. This results in a substantial communication burden, especially when using deep backbones such as ResNet18, whose parameter size is on the order of millions.
In contrast, prototype-based methods including FedProto and FedGH decouple representation learning from parameter aggregation. These methods only exchange class-wise prototypes, and in the case of FedGH, an additional lightweight projection head. As a result, their communication cost is several orders of magnitude lower than that of full-model aggregation methods.
The proposed RefProtoFL follows a hybrid communication strategy. It retains the efficient prototype exchange mechanism while selectively transmitting a sparsified subset of adapter parameters through the Adaptive Parameter Update Dropping (APUD) scheme. The backbone network remains frozen, and only a small portion of task-relevant adapter parameters are aggregated. This design significantly reduces the communication overhead compared to full-model methods, while enabling deeper representation refinement than purely prototype-based approaches.

\subsection{Prototype visualization}
\begin{figure*}[t]
  \centering
  \includegraphics[width=0.9\textwidth]{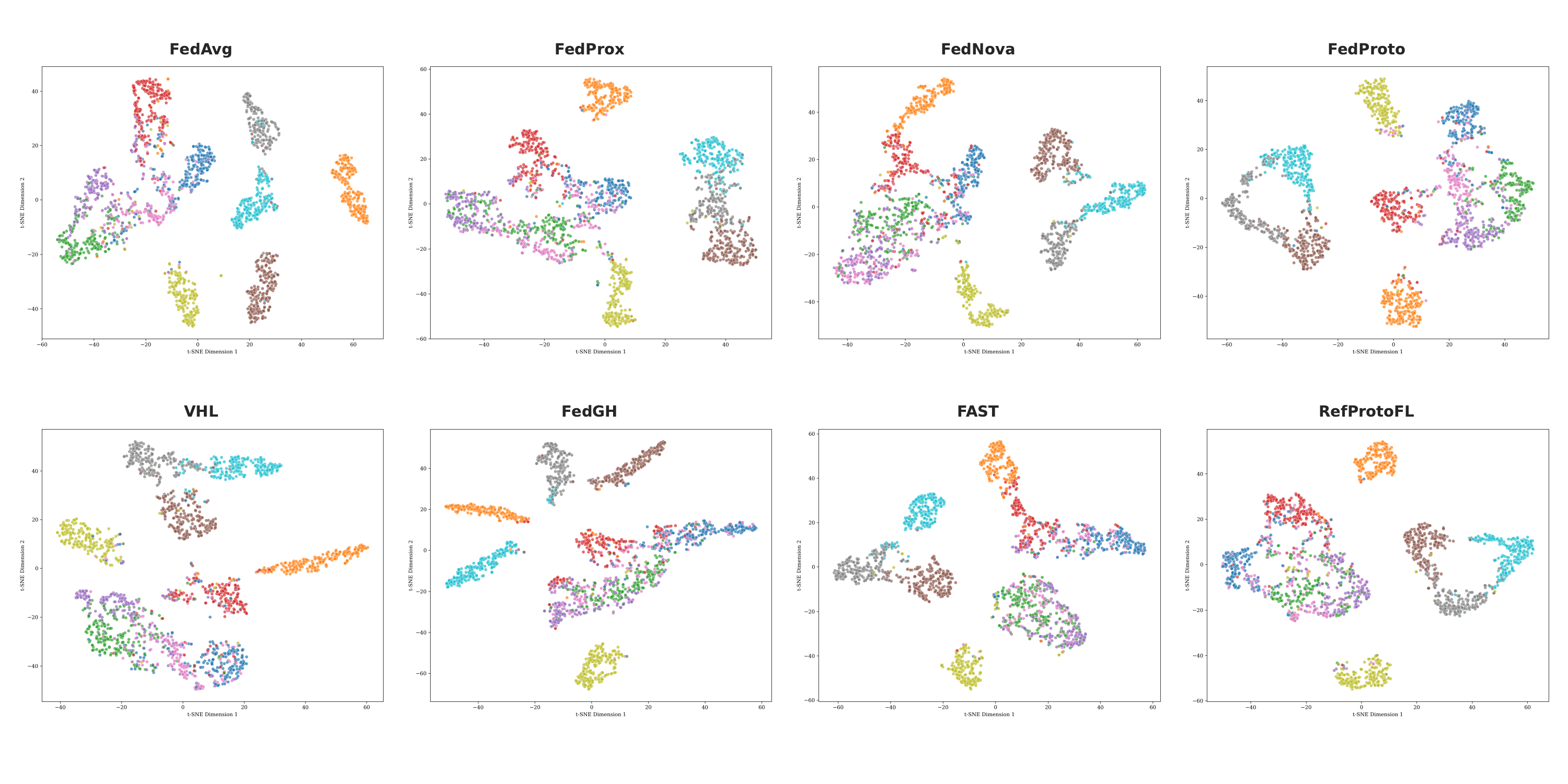}
  \caption{Two-dimensional t-SNE visualization of the learned prototypes of FL methods on the FashionMNIST test set with \(\alpha = 0.5\), where each point represents a sample and colors denote class labels.}
  \label{fig:summary_tsne_combined}
\end{figure*}

Figure \ref{fig:summary_tsne_combined} presents the two-dimensional t-SNE visualization of the learned prototypes on the FashionMNIST test set under the high-heterogeneity setting (\(\alpha=0.5\)). Each point corresponds to a test sample, and different colors indicate different class labels. 

\begin{table}[h]
\centering
\caption{Ablation study of RefProtoFL on CIFAR10 dataset.}
\begin{tabular}{lc}
\toprule
\textbf{Method} & \textbf{Accuracy (\%)} \\
\midrule
RefProtoFL (Full) & \textbf{45.51} \\ 
RefProtoFL (w/o ERPA) & 44.62 \\ 
RefProtoFL (w/o APUD) & 45.37 \\ 
RefProtoFL (w/o APUD \& ERPA) & 44.54 \\
\bottomrule
\end{tabular}
\label{tab:ablation}
\end{table}

\begin{table*}[t]
\centering
\caption{Comparison of different FL methods on Top-1 accuracy.}
\resizebox{\textwidth}{!}{
\begin{tabular}{lccccccccc}
\toprule
\textbf{Dataset} & \textbf{$\alpha$} & \textbf{FedAvg} & \textbf{FedProx} & \textbf{FedNova} & \textbf{FedProto} & \textbf{VHL} & \textbf{FedGH} & \textbf{FAST} & \textbf{RefProtoFL} \\ \midrule
MNIST & 0.5 & 88.18 & 87.41 & 87.87 & 87.68 & 86.83 & 87.29 & 75.19 & \textbf{88.46} \\ 
 & 100 & 97.83 & 96.90 & 97.82 & \textbf{98.33} & 97.08 & 97.87 & 86.74 & 98.25 \\ 
\midrule
FashionMNIST & 0.5 & 73.72 & 70.84 & 73.70 & 73.82 & 72.62 & 73.71 & 63.03 & \textbf{74.21} \\ 
 & 100 & 85.10 & 83.95 & 84.91 & \textbf{85.96} & 84.29 & 85.74 & 76.44 & 85.76 \\ 
\midrule
CIFAR10 & 0.5 & 43.93 & 42.70 & 43.72 & 44.33 & 40.30 & 43.38 & 35.14 & \textbf{45.51} \\ 
 & 100 & 52.05 & 50.37 & 51.73 & 56.05 & 48.84 & 52.94 & 45.07 & \textbf{56.36} \\ 
\midrule
CIFAR100 & 0.5 & 14.56 & 12.81 & 14.46 & 16.28 & 13.24 & 14.16 & 9.97 & \textbf{16.50} \\ 
 & 100 & 16.75 & 8.74 & 17.22 & 18.42 & 15.57 & 14.55 & 12.13 & \textbf{20.02} \\ 
\midrule
\textbf{Average} & - & 59.01 & 56.71 & 58.93 & 60.11 & 57.35 & 58.71 & 50.46 & \textbf{60.63} \\ 
\bottomrule
\end{tabular}
}
\label{tab:full_benchmark}
\end{table*}

\subsection{Ablation Study}
As reported in Table \ref{tab:ablation}, we conduct a systematic ablation study of the proposed RefProtoFL on the CIFAR-10 dataset under a heterogeneous setting with \(\alpha = 0.5\). Overall, the full RefProtoFL achieves the best classification accuracy (45.51\%), which corroborates the effectiveness of jointly integrating the two core components, APUD and ERPA, for performance improvement.

Removing ERPA (w/o ERPA) leads to a notable drop in accuracy to 44.62\%, indicating that the external-referenced prototype alignment mechanism plays a critical role in mitigating class-representation shifts induced by non-IID data distributions. Specifically, ERPA leverages external prototypes constructed from a small public dataset to provide clients with more stable class-level semantic references, thereby enhancing the discriminability of the aggregated global model.

In contrast, disabling APUD alone (w/o APUD) results in only a marginal decrease to 45.37\%. This suggests that the adaptive probabilistic update dropping mechanism primarily contributes to communication efficiency and the stability of parameter selection; its direct impact on the final accuracy is relatively limited, while still offering a consistent positive gain.

When both APUD and ERPA are removed (w/o APUD \& ERPA), the accuracy further decreases to 44.54\%, which is close to the performance observed without ERPA. This observation further highlights ERPA as the primary driver of the accuracy improvements in RefProtoFL, whereas APUD complements the framework by preserving model performance while enabling communication-efficient optimization.

\section{Conclusion}
In this paper, we proposed RefProtoFL, a communication-efficient federated learning framework tailored for edge computing scenarios, where both data heterogeneity and bandwidth constraints pose significant challenges. By combining Adaptive Probabilistic Update Dropping and External-Referenced Prototype Alignment, RefProtoFL effectively reduces communication costs while improving cross-client representation consistency. Extensive simulation-based experiments on multiple image classification benchmarks demonstrate that RefProtoFL consistently outperforms state-of-the-art baselines under severe data heterogeneity, while maintaining competitive performance in near-IID settings.

While the current evaluation is conducted in a simulated federated environment, our long-term goal is to deploy RefProtoFL on real-world edge devices to validate its practical effectiveness under realistic system constraints, such as limited computation resources, unstable connectivity, and heterogeneous hardware. Future work will focus on system-level implementation, real-device experimentation, and extending RefProtoFL to more complex tasks and modalities in real edge federated learning settings.

\section{Acknowledgements}
This work is supported by the National Natural Science Foundation of China grant No. 62032016 and 62372323, and the Foundation of Yunnan Key Laboratory of Service Computing under Grant YNSC24105. 

\bibliographystyle{named}
\bibliography{ijcai26}

\end{document}